\documentclass[journal]{new-aiaa}

\usepackage{subcaption}
\usepackage{graphicx}
\usepackage{amsmath, bbm}
\usepackage[version=4]{mhchem}
\usepackage{siunitx}
\usepackage{longtable,tabularx}
\usepackage{breqn}
\usepackage{svg}
\usepackage{placeins}

\usepackage{cleveref}

\crefname{equation}{}{}

\usepackage{xcolor} 

\newtheorem{helptheorem}{Theorem}{} 
 
\newtheorem{helplemma}[helptheorem]{Lemma} 
\newtheorem{helpcorollary}[helptheorem]{Corollary} 
 
\newtheorem{helpexample}[helptheorem]{Example} 
 
\newtheorem{helpproposition}[helptheorem]{Proposition} 
 
\newtheorem{helpremark}[helptheorem]{Remark} 
 
\newtheorem{helpdefinition}[helptheorem]{Definition}
 
\newtheorem{helpassumption}[helptheorem]{Assumption}

\newtheorem{helpproblem}[helptheorem]{Problem}

\newtheorem{fact}{\indent Fact}

\begin{document}

\title{Stability Analysis of Deep Reinforcement Learning for Multi-Agent Inspection in a Terrestrial Testbed\footnote{The views expressed are those of the author and do not necessarily reflect the official policy or position of the Department of the Air Force, the Department of Defense, or the U.S. government. Approved for public release; distribution is unlimited. Public Affairs approval \#AFRL-2024-6672.}}

\author{Henry Lei\footnote{henry.lei@verusresearch.net}, Zachary S. Lippay, Anonto Zaman, Joshua Aurand\footnote{joshua.aurand@verusresearch.net}, and Amin Maghareh}
\affil{Verus Research, Albuquerque, NM, 87110}

\author{Sean Phillips}
\affil{Air Force Research Laboratory, Kirtland AFB, NM, 87117}
\date{November 2024}

\maketitle

\begin{abstract}
The design and deployment of autonomous systems for space missions require robust solutions to navigate strict reliability constraints, extended operational durations, and communication challenges. This study evaluates the stability and performance of a hierarchical deep reinforcement learning (DRL) framework designed for multi-agent satellite inspection tasks. The proposed framework integrates a high-level guidance policy with a low-level motion controller, enabling scalable task allocation and efficient trajectory execution. Experiments conducted on the Local Intelligent Network of Collaborative Satellites (LINCS) testbed assess the framework’s performance under varying levels of fidelity, from simulated environments to a cyber-physical testbed. Key metrics, including task completion rate, distance traveled, and fuel consumption, highlight the framework's robustness and adaptability despite real-world uncertainties such as sensor noise, dynamic perturbations, and runtime assurance (RTA) constraints. The results demonstrate that the hierarchical controller effectively bridges the sim-to-real gap, maintaining high task completion rates while adapting to the complexities of real-world environments. These findings validate the framework’s potential for enabling autonomous satellite operations in future space missions.
\end{abstract}

\section*{Nomenclature}
\label{section:Nomenclature}
{\renewcommand\arraystretch{1.0}
\noindent\begin{longtable*}{@{}l @{\quad=\quad} l@{}}
$A$ & Set of Inspection Points in the Hill's frame \\
$a_c$ & Maximum allowable acceleration\\
$b_i$ & Barrier function for velocity level constraints\\
$F_E$ & Earth-Centered Inertial Frame\\
$F_H$ & Hill's Frame\\
$f_c$ & Thrust limits in the Hill's frame\\
$\mathbf{f}_{\text{Hill}}$ & Analytical Solution for Inspector Relative Position\\
$\mathbf{f}_{e,i}$ & External Applied Forces\\
$\mathbf{f}_{p,i}$ & Environment Perturbation Forces\\
$h_{ij}$ & Barrier Function for position level constraints\\
$\mathcal{I}_d$ & Set of Inspection Satellite Id's\\
$\mathbb{I}_{\cdot}$ & Indicator function taking the value 1 on prespecified conditions\\
$J_2$ & J2 Coefficient\\
$m_i$ & Satellite Mass\\
$n$ & Target RSO Mean Motion Parameter\\

$\mathbf{O}_{ll}$ & Observation function for Single Agent Point to Point Controller\\
$\mathbf{O}_{hl}$ & Observation function for High Level Planner\\
$R_E$ & Radius of the Earth\\
$\textrm{rew}_{ll}$ & Reward function for Single Agent Point to Point Controller\\
$\textrm{rew}_{hl}$ & Reward function for High Level Planner\\
$r_c$ & Minimum allowable distance\\
$\mathbf{r}_i$ & Inspection Satellite Position\\
$\delta\mathbf{r}_i$ & Relative Position of Inspection Satellite\\
$\mathcal{U}$ & Uniform distribution\\
$\mathbf{u}_i(t)$ & Inspection Satellite Thrust Force\\
$v_c$ & Maximum allowable velocity\\
$\mu$ & Earth Gravitational Parameter\\

\end{longtable*}}
\section{Introduction}
\label{section:Introduction}
The design of autonomous systems for space missions poses significant challenges due to strict reliability requirements, extended operational lifespans, and the inherent difficulty of maintaining consistent ground communication, particularly in high-altitude orbits. As a result, there has been a growing emphasis on incorporating advanced autonomous capabilities into spacecraft. While traditional control-theoretic methods have been widely utilized to enable basic autonomy in satellites, recent advancements in machine learning (ML) and cognitive AI have sparked considerable interest within the research community. These technologies show immense promise in addressing complex decision-making problems across various domains~\cite{mnih2015human, lei2022deep}. However, several critical challenges continue to hinder their widespread adoption, particularly in reinforcement learning (RL)-based policies.

One of the primary limitations is the ``black-box'' nature of RL models, which makes their decision-making processes difficult to interpret and guarantee. This opacity raises significant concerns, as minor disturbances---such as a single-pixel alteration in input images---can lead to catastrophic instabilities, a well-documented problem in robust RL research~\cite{goodfellow2015explaining, zhang2021robust}. Additionally, deploying RL-trained policies in new environments, particularly transitioning from simulations to physical hardware, often results in significant performance degradation due to cumulative discrepancies---a challenge referred to as the ``sim-to-real gap''~\cite{tobin2017domain}. Bridging this gap is a key focus of ongoing research in transfer learning and domain adaptation~\cite{pan2010survey}. The topic of robustness has been extensively investigated in optimization theory; therein an underlying uncertainty set is typically specified - under which the optimizer must manage all possible deviations induced by such uncertainty. Since the RL agents continuously observes system response through interaction, their ability to effectively sample low-probability outcomes within the uncertainty set is limited. This difficulty is exaggerated when access to real and/or representative system data is limited (such as in the space environment). More details regarding this can be found in~\cite{morimoto2005, pinto2017, tamar2013, tessler2019}. Although such difficulties exist, there are some key advantages to training on simulated data. For instance, this commonly allows for accelerated data collection through parallelization, safe experimentation with failure scenarios, and access to extreme conditions that are impractical or impossible to replicate in the real world---such as those encountered during long-duration space missions~\cite{kalakrishnan2013learning, nguyen2020deep}. To take advantage of this in a meaningful way, it is crucial to analyze how the byproduct (a trained RL control policy) will perform in real-world environments beyond the scope implied by simulation. To do so, we leverage the LINCS lab, which offers a tested platform for addressing such challenges. By utilizing a terrestrial testbed with quadrotors emulating satellite on-orbit dynamics, LINCS enables detailed evaluations of policy performance under controlled experimental conditions. This facility provides the infrastructure and hardware necessary for collecting trajectory and performance data across simulated and physical scenarios~\cite{phillips2024lincs}.

In this study, we evaluate the stability and performance of a model hierarchical inspection controller~\cite{lei2022deep} under varying model uncertainties and dynamic perturbations using the LINCS cyber-physical testbed. While \cite{lei2022deep} studied the efficacy of an RL policy trained to solve a vehicle routing problem, this work investigates the robustness of the control scheme against different types of unknown and unmodeled uncertainties. The advantage of using \cite{lei2022deep} is the relative simplicity of the environmental formulation used for training. This helps streamline single-factor attribution during experimentation. A baseline performance metric is first established in the training environment, followed by testing in a minimally altered but realistic scenario representing the complete inspection problem. Experiments are conducted to isolate and analyze the effects of modeling uncertainties and perturbations on key performance metrics, including task completion rate, time taken, and distance traveled. Although more formal testing is needed to draw statistically significant conclusions, we find that the hierarchical structure helps preserve task completion rate despite significant impacts to time taken and distance traveled. This analysis helps provide critical insight into the robustness and adaptability of autonomous policies in complex operational contexts.

A more detailed exposition of the problem formulation and key pieces of background information are introduced below in Section~\ref{section:problem_formulation}. This is followed by experimental setup and testing methodology; see Section~\ref{section:experimental_setup}. The results are then introduced in Section~\ref{section:expr_results} with concluding remarks saved for Section~\ref{section:conclusion}.

\section{Problem Formulation}
\label{section:problem_formulation}
In this work, we analyze the effects of environmental changes on the performance of a pre-trained DRL policy in a multi-agent satellite inspection scenario. The problem under consideration in Section~\ref{subsec: ProblemForm} is defined through the following prerequisite elements: key reference frames \ref{subsec: frame def}, vehicle dynamics \ref{subsec: 2body}-\ref{subsec: CWH}, and runtime assurance formulation \ref{subsec: RTA}; all described below.

\subsection{Frame Definitions}\label{subsec: frame def}
The Earth-Centered Inertial (ECI) frame, denoted as $F_E$, is defined by the orthogonal unit vectors $\mathbf{i}_E$, $\mathbf{j}_E$, and $\mathbf{k}_E$, with origin at $O_E$. The $\mathbf{i}_E$ direction points toward the vernal equinox, $\mathbf{k}_E$ aligns with Earth's north celestial pole, and $\mathbf{j}_E$ completes the right-handed coordinate system.

For $n \in \mathbb{N}$ deputy spacecraft indexed by $\mathcal{I}_d = \{1, 2, \dots, n\}$ and a chief spacecraft (alternatively referred to as an RSO) indexed as $i = 0$, the full agent set is $\mathcal{I} = \mathcal{I}_d \cup \{0\}$. Each spacecraft $i \in \mathcal{I}$ has a body-fixed frame $F_i$, defined by orthogonal unit vectors $\mathbf{i}_i$, $\mathbf{j}_i$, and $\mathbf{k}_i$, with origin at $O_i$.

The Hill's frame, $F_H$, is fixed to the chief spacecraft, with origin $O_H = O_0$. Its axes are defined as follows: $\mathbf{i}_H$ points radially from $O_0$ toward $O_E$, $\mathbf{k}_H$ aligns with the chief's angular momentum vector, and $\mathbf{j}_H$ completes the right-handed system.

\subsection{Two-Body Dynamics}\label{subsec: 2body}
The position of spacecraft $i \in \mathcal{I}$ relative to $O_E$ is denoted by $\mathbf{r}_i \in \mathbb{R}^3$. The dynamics of spacecraft $i$ in the ECI frame are:
\begin{equation}
    \ddot{\mathbf{r}}_i(t) = -\frac{\mu}{\|\mathbf{r}_i(t)\|^3}\mathbf{r}_i(t) + \frac{1}{m_i}\big(\mathbf{f}_{p,i}(t) + \mathbf{f}_{e,i}(t)\big),
\end{equation}
where $\mu$ is Earth's gravitational parameter, $m_i$ is the spacecraft mass, $\mathbf{f}_{p,i}$ represents environmental perturbation forces, and $\mathbf{f}_{e,i}$ is the external force. For this study, only the J2 perturbation is considered. The perturbation force is modeled as:
\begin{equation}
    \mathbf{f}_{p,i}(t) = -m_i J_2 \frac{3\mu}{2\|\mathbf{r}_i(t)\|^5} \left(\frac{R_E}{\|\mathbf{r}_i(t)\|}\right)^2
    \begin{bmatrix}
        x_i\left(1 - 5z_i^2 / \|\mathbf{r}_i(t)\|^2\right) \\
        y_i\left(1 - 5z_i^2 / \|\mathbf{r}_i(t)\|^2\right) \\
        z_i\left(3 - 5z_i^2 / \|\mathbf{r}_i(t)\|^2\right)
    \end{bmatrix},
\end{equation}
where $J_2$ is the coefficient of Earth's oblateness, $R_E$ is Earth's mean radius, and $\mathbf{r}_i(t) = [x_i, y_i, z_i]^\top$. 



\subsection{Linearized Clohessy-Wiltshire Dynamics in Hill’s Frame}\label{subsec: CWH}

For a deputy spacecraft $i \in \mathcal{I}_d$, the relative position vector with respect to the chief spacecraft is defined as $\delta\mathbf{r}_i = \mathbf{r}_i - \mathbf{r}_0$. Resolved in Hill's frame $F_H$, $\delta\mathbf{r}_i \in \mathbb{R}^3$ describes the deputy’s relative motion with respect to the chief. To linearize the dynamics, the following assumptions are made:
\begin{enumerate}[label=(A\arabic*)]
    \item The chief spacecraft is translationally unactuated.
    \item The chief spacecraft is in a circular orbit.
    \item The relative distances between deputies and the chief are sufficiently small, satisfying $\|\delta\mathbf{r}_i\| \ll \|\mathbf{r}_0\|$.
    \item The relative positions and velocities of deputies are known, ensuring consensus on the chief's state.
\end{enumerate}

Under these assumptions, the linearized dynamics of the $i$th deputy in Hill's frame can be expressed as:
\begin{equation}\label{eqn: Hill Dynamics}
    \begin{bmatrix}
        \dot{\delta\mathbf{r}}_i(t) \\
        \ddot{\delta\mathbf{r}}_i(t)
    \end{bmatrix}
    =
    \begin{bmatrix}
        \mathbf{0}_{3 \times 3} & \mathbf{I}_{3 \times 3} \\
        \mathbf{A} & \mathbf{B}
    \end{bmatrix}
    \begin{bmatrix}
        \delta\mathbf{r}_i(t) \\
        \dot{\delta\mathbf{r}}_i(t)
    \end{bmatrix}
    +
    \begin{bmatrix}
        \mathbf{0}_{3 \times 3} \\
        \frac{1}{m_i}\mathbf{I}_{3 \times 3}
    \end{bmatrix}
    \mathbf{u}_i(t),
\end{equation}
where:
\begin{align}
    \mathbf{A} &= 
    \begin{bmatrix}
        3n^2 & 0 & 0 \\
        0 & 0 & 0 \\
        0 & 0 & -n^2
    \end{bmatrix},
    \quad
    \mathbf{B} = 
    \begin{bmatrix}
        0 & 2n & 0 \\
        -2n & 0 & 0 \\
        0 & 0 & 0
    \end{bmatrix}.
\end{align}

Here, $n = \sqrt{\mu / \|\mathbf{r}_0\|^3}$ is the mean motion of the chief spacecraft and $\mathbf{u}_i(t)$ is the control force applied in $F_H$. $\mathbf{I}_{3 \times 3}$ is the $3 \times 3$ identity matrix, and $\mathbf{0}_{3 \times 3}$ is the $3 \times 3$ zero matrix. The force $\mathbf{u}_i(t)$ is resolved in Hill's frame but can be transformed into the inertial frame $F_E$ using the rotation matrix $\mathbf{R}_{E/H}$:
\begin{equation}
    \mathbf{f}_{c,i} = \mathbf{R}_{E/H} \mathbf{u}_i(t).
\end{equation}
The relative motion $\delta\mathbf{r}_i(t)$ can be obtained by forward-integrating the linearized dynamics:
\begin{equation}
    \delta\mathbf{r}_i(t) = \mathbf{f}_{\text{Hill}}(t, \mathbf{u}_i(t)) := \delta\mathbf{r}_i(0) + \int_0^t \dot{\delta\mathbf{r}}_i(s) \, ds,
\end{equation}
where $\dot{\delta\mathbf{r}}_i(s)$ satisfies the linearized equation of motion specified in eqn.~\eqref{eqn: Hill Dynamics}.

\subsection{Runtime Assurance}\label{subsec: RTA}
To ensure safe and efficient operations of the spacecraft during trajectory planning and execution, runtime assurance is implemented. This involves enforcing constraints on position, velocity, acceleration, and control inputs. A relaxed quadratic program is utilized to guarantee feasibility while maintaining optimal performance. Key results related to control barrier functions and higher-order control barrier functions can be found in \cite{hocbf_Xiao_22}.

\subsubsection{Position Constraint}
Collision avoidance between spacecraft is ensured by maintaining a minimum distance between agents. For a pair of spacecraft $i, j \in \mathcal{I}_d$ such that $i \neq j$, the position constraint is formulated using a barrier function:
\begin{equation}
    h_{ij}(\delta\mathbf{r}_i, \delta\mathbf{r}_j) = \frac{1}{2} \left(\|\delta\mathbf{r}_i - \delta\mathbf{r}_j\|^2 - r_c^2\right),
\end{equation}
where $r_c > 0$ is the minimum allowable distance. To ensure this constraint is met, the first and second time derivatives of $h_{ij}$ along the relative trajectory are used to obtain a higher-order constraint. This is used to maintain safety over time, with the derived constraint becoming:
\begin{equation}\label{eqn: pos constraint}
    \ddot{h}_{ij} \left( \delta \mathbf{r}_i, \delta \mathbf{r}_j, \dot{\delta \mathbf{r}}_i, \dot{\delta \mathbf{r}}_j, \ddot{\delta \mathbf{r}}_j, \mathbf{u}_i \right) + (\gamma_1 + \gamma_0)\dot{h}_{ij} \left( \delta \mathbf{r}_i, \delta \mathbf{r}_j, \dot{\delta \mathbf{r}}_i, \dot{\delta \mathbf{r}}_j \right) + \gamma_1\gamma_0 h_{ij} \left( \delta \mathbf{r}_i, \delta \mathbf{r}_j \right) \geq 0,
\end{equation}
where $\gamma_0, \gamma_1 > 0$ are control parameters to adjust the responsiveness of the barrier function. Without loss of generality, this applies to the chief for $j=0$ giving $\delta \mathbf{r}_j = \dot{\delta \mathbf{r}}_j = \ddot{\delta \mathbf{r}}_j = 0$ in eqn.~\eqref{eqn: pos constraint}.

\subsubsection{Velocity Constraint}
The relative velocity of each deputy spacecraft is constrained to stay within an allowable upper bound. For spacecraft $i \in \mathcal{I}_d$, the velocity constraint is defined by:
\begin{equation}
    b_i(\dot{ \delta \mathbf{r}}_i) = \frac{1}{2} \left(v_c^2 - \|\dot{ \delta \mathbf{r}}_i\|^2\right),
\end{equation}
where $v_c > 0$ is the maximum allowable velocity. 
To ensure the velocity constraint holds, we take the time derivative of $b_i$ along its trajectory and derive the following constraint:
\begin{equation}
    \dot{b}_i\left( \delta \mathbf{r}_i, \dot{ \delta \mathbf{r}}_i, \mathbf{u}_i \right) + \gamma_2 b_i\left( \dot{ \delta \mathbf{r}}_i \right) \geq 0,
\end{equation}
where $\gamma_2 > 0$ is a control parameter.

\subsubsection{Acceleration and Input Constraints}
The spacecraft \textit{relative} acceleration is constrained by an upper bound $a_c > 0$:
\begin{equation}
    a_c^2 -  \left( \ddot{ \delta \mathbf{r}}_i \right)^{\rm T} \left( \mathbf{A} \delta \mathbf{r}_i + \mathbf{B}\dot{\delta  \mathbf{r}}_i + \frac{\mathbf{u}_i}{m_i}\right)  \geq 0.
\end{equation}
Note that $\ddot{ \delta \mathbf{r}}_i$ is obtained as the measurement of the $i$th agent's acceleration calculated in $F_H$. Thrust inputs are constrained by upper bounds on each component:
\begin{equation}
    u_c^{(k)} - |\mathbf{u}_i^{(k)}| \geq 0 , \quad k = 1, 2, 3
\end{equation}
where $u_c^{(k)} > 0$ represents the maximum allowable thrust along each axis.

\subsubsection{Relaxed Quadratic Program}
To satisfy the constraints while ensuring feasibility, a relaxed quadratic program (QP) is formulated. Slack variables $\boldsymbol{\phi}_i$ are introduced to soften the constraints. The QP for spacecraft $i \in \mathcal{I}_d$ is given as:
\begin{equation}
    (\mathbf{u}_i^*, \boldsymbol{\phi}_i^*) = \underset{\mathbf{u}_i, \boldsymbol{\phi}_i}{\text{arg~min}} \|\mathbf{u}_i - \mathbf{a}_i\|^2 + \gamma_3 \|\boldsymbol{\phi}_i\|^2,
\end{equation}
subject to:
\begin{align}
    \ddot{h}_{ij} + (\gamma_1 + \gamma_0)\dot{h}_{ij} + \gamma_1\gamma_0 h_{ij} &\geq \phi_{ij}, \quad \forall j \neq i, \\
    \dot{b}_i + \gamma_2 b_i &\geq \phi_i^{(v)}, \\
    a_c^2 - \left( \ddot{\delta\mathbf{r}}_i \right)^{\rm T} \left( \mathbf{A} \delta \mathbf{r}_i + \mathbf{B} \dot{\delta \mathbf{r}}_i + \frac{\mathbf{u}_i}{m_i}\right)  &\geq  \phi_i^{(a)}, \\
    u_c^{(k)} - |\mathbf{u}_i^{(k)}| &\geq  \phi_i^{(u)}, \quad k = 1, 2, 3.
\end{align}

Here, $\mathbf{a}_i$ represents the desired nominal control, $\gamma_3 > 0$ is a penalty weight for slack variables, and $\boldsymbol{\phi}_i = [\phi_{ij}, \phi_i^{(v)}, \phi_i^{(a)}, \phi_i^{(u)}]$ is the vector of slack variables. The objective minimizes the deviation from the nominal control while penalizing constraint violations. This formulation ensures that all constraints are respected, and feasibility is maintained even under challenging dynamic conditions.

\subsection{Inspection Mission and Hierarchical Control}\label{subsec: ProblemForm}

The inspection mission involves coordinating multiple agents to autonomously inspect predefined points while minimizing resource consumption and maintaining safety constraints. A hierarchical control framework is employed, consisting of:
\begin{enumerate}
    \item A \textbf{high-level planner}, which determines the sequence of inspection points for each agent.
    \item A \textbf{low-level controller}, which generates control inputs to drive agents to goal waypoints.
\end{enumerate}

The problem is modeled as a multi-agent Markov Decision Process (MDP), where each agent operates within an environment defined by its state, action space, and a shared reward function. The hierarchical decomposition leads to two sub-problems: guidance (Problem 1) and motion control (Problem 2).

\subsubsection*{Problem 1: Guidance Problem}
The guidance problem determines the optimal sequence of inspection points for each agent to minimize total travel distance while ensuring all points are visited. Let \( A = \{a_1, a_2, \dots, a_m\} \) denote the set of inspection points, and let \( J_i \) represent the ordered sequence of points assigned to agent \( i \in \mathcal{I}_d \). The optimization problem is defined through the calculation of:

\begin{equation}
    \mathbf{a}^* = \arg \min_{\mathbf{a}} \sum_{i \in \mathcal{I}_d} \sum_{j=1}^{|J_i|} \|\mathbf{a}_{i,j+1} - \mathbf{a}_{i,j}\|,
\end{equation}
subject to:
\begin{equation}
    \left|\bigcup_{i \in \mathcal{I}_d} J_i\right| = m,
\end{equation}
where \( \mathbf{a}_{i,j} \in A \) is the \( j \)-th inspection point in agent \( i \)'s sequence. The cardinality constraint ensures that all points in \( A \) are assigned to at least one agent.

The guidance problem focuses on determining the global strategy for allocating inspection tasks among agents. The optimization objective minimizes the cumulative travel distance across all agents while ensuring complete coverage of the inspection area. The solution provides a sequence of waypoints for each agent, which serve as inputs to the motion control problem.

\subsubsection*{Problem 2: Motion Control Problem}
The motion control problem involves generating a control policy for each agent to move between assigned waypoints efficiently while adhering to safety and performance constraints. For agent \( i \in \mathcal{I}_d \), the optimization problem is defined through the calculation of:

\begin{equation}\label{eqn: LL_control}
    \mathbf{u}_i^* = \arg \max_{\mathbf{u}_i(t)} \int_0^T \frac{1}{\|\delta\mathbf{r}_i(t) - \delta\mathbf{r}_{g}\| + 1} \, dt,
\end{equation}
subject to:
\begin{equation}
    \delta\mathbf{r}_i(T) = \delta\mathbf{r}_g, \quad \forall i \in \mathcal{I}_d,
\end{equation}
\begin{equation}
    \delta\mathbf{r}_i(t) = \mathbf{f}_{\text{Hill}}(t,\mathbf{u}_i(t)), \quad \forall t \in [0, T],
\end{equation}
\begin{equation}
    \|\mathbf{u}_i(t)\|_{\infty} \leq 1, \quad \forall t \in [0, T].
\end{equation}

The motion control problem ensures each agent can traverse between consecutive waypoints efficiently and safely. The objective maximizes a reward function inversely proportional to the distance to the target, encouraging proximity to the goal. The constraints enforce reaching the goal location $\delta\mathbf{r}_{g}$, bounded control inputs, and the linearized relative motion dynamics in Hill's frame. The solution, \( \mathbf{u}_i^* \), provides the control inputs required to execute point-to-point motion control.

The hierarchical approach integrates the solutions of Problems 1 and 2. The guidance problem assigns global inspection tasks, producing a sequence of waypoints. The motion control problem then translates these waypoints into control inputs for each agent. Together, this framework enables scalable, multi-agent inspection while addressing both global task allocation and local trajectory execution.

\section*{Experimental Setup}
\label{section:experimental_setup}

This section describes the control schemes being tested and the experimental structure used to assess them. There are two distinct controllers that are bridged hierarchically; low-level control - see Section~\ref{subsec: LLControl} and high-level guidance - see Section~\ref{subsec: HLControl}. The testbed used for experiments is described in Section~\ref{subsec: LINCS}. Lastly, the methodology and structure used to evaluate policy stability is described in Section~\ref{subsec: Factors}.

\subsubsection{Single-Agent Point-to-Point Controller}\label{subsec: LLControl}

The single-agent point-to-point control experiment evaluates the performance of a trained neural network controller (NNC) in a hardware-in-the-loop (HIL) satellite emulation environment. Since this setup involves only one agent (\(|\mathcal{I}_d| = 1\)), the agent subscript is omitted for clarity. The objective is to test the controller's ability to navigate from a randomly initialized position to a randomly generated target within a predefined time limit.

\paragraph{Training Environment}
The satellite follows the linearized Clohessy-Wiltshire (CW) dynamics (refer to Section II-D). It is equipped with axially aligned continuous thrusters, each with a maximum thrust of 1 N. At every discrete time step (\(\Delta t = 1\) s), the agent receives a state observation containing distance and direction of $\mathbf{r}_g$ to the deputy in conjunction with a relative velocity measurement. This can be seen below:
\begin{equation}
    \mathbf{O}_{ll} = \left( \frac{\delta \mathbf{r} - \delta \mathbf{r}_g}{1000}, \delta \dot{\mathbf{r}} \right).
\end{equation}
The episode terminates under the following conditions:
\begin{enumerate}
    \item The agent successfully reaches the target, defined as \(\|\delta \mathbf{r} - \delta \mathbf{r}_g\| < 10 \, \text{m}\),
    \item The agent drifts out of bounds,
    \item The episode duration exceeds 500 s.
\end{enumerate}
Training instantiation is rescaled according to dimensions provided by the LINCS aviary where each instance is initialized with a randomized agent state drawn according to 
\begin{equation}
    \delta \mathbf{r}(0) = [1.17, 2.5, 1]\cdot\mathbf{x}, \quad \delta \mathbf{r}_g = [1.17, 2.5, 1]\cdot \mathbf{y},
\end{equation}
where $\mathbf{x}, \mathbf{y} \sim \mathcal{U}([-240m, 240m]^3)$.

\paragraph{Reward Function}
The reward function used for point-to-point control is designed to minimize distance to a target waypoint $\delta \mathbf{r}_g$ in conjunction with the objective in eqn.~\eqref{eqn: LL_control}. The function is conditioned on positional state $\delta \mathbf{r}_{-1}$ and an implicitly defined thrust action $\mathbf{u}$. The effect of $\mathbf{u}$ on the reward $\textrm{rew}(\cdot)$ is encoded by the terms $\delta \mathbf{r}$ and $\delta \dot{\mathbf{r}}$ which reflect vehicle state transition under $\mathbf{u}$ for a single time step. Inspired by the Langrangian associated with eqn.~\eqref{eqn: LL_control}, the deputy is rewarded at each time step based on the following:
\begin{equation}
    \textrm{rew}_{ll}(\delta \mathbf{r}, \delta \mathbf{r}_{-1}, \delta \dot{\mathbf{r}}, \delta \mathbf{r}_g) = 
    \frac{\alpha}{\|\delta \mathbf{r} - \delta \mathbf{r}_g\| + 1} 
    + \beta \left( \|\delta \mathbf{r}_{-1} - \delta \mathbf{r}_g\| - \|\delta \mathbf{r} - \delta \mathbf{r}_g\| \right) 
    - \nu \|\delta \dot{\mathbf{r}}\|_1 \mathbb{I}_{\{\|\delta \dot{\mathbf{r}}\| > \eta \sigma_\mu \|\delta \mathbf{r} - \delta \mathbf{r}_g\| \}}
\end{equation}
where \(\alpha, \beta, \nu\) are weighting factors and \(\eta, \sigma_\mu\) defines the maximum allowable velocity threshold. The first and second terms incentivize distance reduction to a goal waypoint with additional reinforcement provided to movement along a straight line. The third term encodes a variable speed limit penalizing conditions that may be unsafe or difficult to control.

\subsubsection{Multiagent Guidance Planner}\label{subsec: HLControl}

The multiagent guidance planner operates as a high-level control policy for allocating and sequencing inspection tasks across multiple agents. The environment is modeled as a vehicle routing problem on a fixed inspection graph, consisting of 20 predefined inspection points in the CWH frame. These points form an ellipsoid centered around the inspection target, scaled to match the dimensions of the LINCS aviary, with a nominal radius of 300 meters. For an example of the graph and overlayed trajectories, pleass see Figure~\ref{fig:hier_rollout} below.

\paragraph{Training Environment}
The training environment uses a centralized policy network to manage all agents in the inspection team, where \(|\mathcal{I}_d| = 3\). At each environment step, the network selects an action \(\mathbf{a}_i \in A\) for each agent \(i \in \mathcal{I}_d\), based on the following observation:
\begin{equation}
    \mathbf{O}_{hl} = (\mathbf{p}, \mathbf{v}),
\end{equation}
where:
\begin{itemize}
    \item \(\mathbf{p} \in \mathbb{R}^3\): The index position of agent \(i\) for $i=1,\dots, 3$ in the inspection graph $A$,
    \item \(\mathbf{v} \in \{0, 1\}^{20}\): A Boolean vector indicating whether each inspection point has been visited.
\end{itemize}
A centralized policy network is implemented where experience sharing can be effectively leveraged to improve efficacy of optimization between environment steps. Each episode is instantiated according to a uniformly drawn initial state for agent position. ie. $\delta\mathbf{r} \in \mathcal{U}(A^{3})$. The episode terminates once all points in $A$ have been visited.

\paragraph{Reward Function}
The agents collectively receive a shared reward signal to encourage efficient task completion while penalizing redundant or conflicting actions. For an agent factored, joint action $\mathbf{a} = (a_{1},a_{2},a_{3})$ and corresponding indices $\mathbf{p} = (p_{1},p_{2},p_{3})$ the shared reward function is defined as:
\begin{equation}
    \textrm{rew}_{hl}(\mathbf{a}, \mathbf{v}_{-1}, \mathbf{a}_{-1}) = -\sum_{i=1}^{3}\left(\alpha \|a_{i}-a_{i,-1}\| +  \beta\, \mathbb{I}_{p_{i}\in\mathbf{v}_{-1}} + \nu\,\mathbb{I}_{p_{i} = p_{k},\,k\neq i}\right).
\end{equation}
This formulation encourages agents to minimize cumulative travel distance while simultaneously avoiding point revisitation and scenarios resulting in a high probability of collision.  

\subsubsection{LINCS Lab}\label{subsec: LINCS}

The Local Intelligent Network of Collaborative Satellites (LINCS) Lab serves as a testing facility for on-orbit satellite operations. It provides a comprehensive platform for both simulation-based and hardware-in-the-loop (HIL) experiments. The lab’s infrastructure is built on the Robot Operating System 2 (ROS2) communication architecture, supporting flexible and scalable testing scenarios. For more detail please see \cite{phillips2024lincs}.

\paragraph{Simulation Environment}
The LINCS simulation environment consists of two key components:
\begin{enumerate}
    \item \textbf{Satellite Dynamics:} Translational motion is modeled using two-body dynamics with J2 perturbations, while attitude dynamics are governed by the rigid-body equations.
    \item \textbf{Flight Software:} Guidance and control commands are generated using the hierarchical DRL algorithm, ensuring optimized trajectory planning and constraint enforcement. Translational commands are validated through real-time RTA to maintain state constraints.
\end{enumerate}

\begin{figure}[!htb]
\begin{subfigure}[b]{.45\textwidth}
    \centering
    \includegraphics[width=\textwidth]{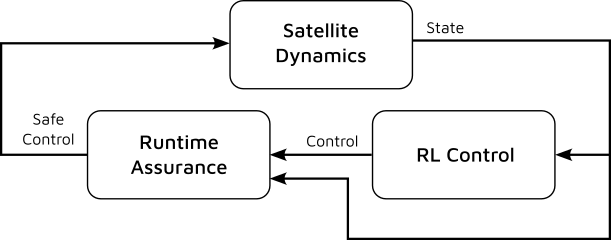}
\end{subfigure}
\hfill
\begin{subfigure}[b]{0.45\textwidth}
\centering
\includegraphics[width=\textwidth]{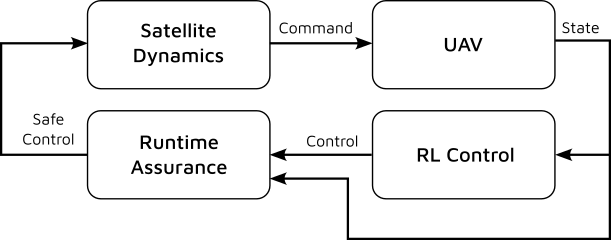}
\end{subfigure}
\caption{The left diagram shows simulation workflow with RTA enforcement in the LINCS Lab. The right diagram shows an analogous version for LINCS Cyber-Physical emulation using quadrotor UAVs.}
\label{fig:LINCS_block_diagram}
\end{figure}

Figure~\ref{fig:LINCS_block_diagram} illustrates the simulation workflow, where the RTA ensures the satellite adheres to performance and safety limits. Key RTA parameters are listed in Table~\ref{tab:rta_parameters}.

\paragraph{Cyber-Physical Emulation}
In addition to simulation capability, the LINCS Lab provides a platform for cyber-physical emulation using quadrotor UAVs to replicate satellite dynamics. The UAVs are tracked in real time using a Vicon motion capture system, which communicates positional data at 50 Hz. The key features of the emulation setup include:
\begin{itemize}
    \item Wireless reference trajectory transmission to the UAV via Raspberry Pi 4,
    \item Onboard computation of feedback and trajectory commands using Dronekit APIs,
    \item Stabilization using ArduCopter 4.4 for attitude and altitude control.
\end{itemize}
The full process is shown in Figure~\ref{fig:LINCS_block_diagram}.

\paragraph{Mixed Simulation and Emulation}
The LINCS Lab also supports mixed setups, where agents operate in both simulation and emulation environments simultaneously. For example, in a scenario with \(n=3\) agents, two agents may operate in simulation while the third is controlled via hardware. This mixed setup enables testing of system scalability and robustness through fine-tuned experimental control over a variety of test factors. Key parameters used in the LINCS Lab experiments, including collision avoidance radius, maximum velocity, acceleration, and thrust limits, are detailed in Table~\ref{tab:rta_parameters}.

\begin{table}[h!]
    \centering
    \begin{tabular}{|c|c|}
        \hline
        \textbf{Parameter} & \textbf{Value} \\ \hline
        Collision Radius (\(r_c\)) & 50 m \\ \hline
        Maximum Velocity (\(v_c\)) & 3 m/s \\ \hline
        Maximum Acceleration (\(a_c\)) & 1.732 m/s\(^2\) \\ \hline
        Input Upper Bound (\(f_c\)) & 1 N \\ \hline
    \end{tabular}
    \caption{RTA Parameters for LINCS Experiments}
    \label{tab:rta_parameters}
\end{table}
\vspace{-10pt}
\subsubsection{Experimental Setup}\label{subsec: Factors}
DRL policies are often trained in environments that differ significantly from their deployment conditions. This disparity, commonly referred to as the "sim-to-real gap," arises from the trade-off between simulation fidelity and computational cost. Higher-fidelity simulations require greater resources to develop and train, while simpler environments often fail to capture the complexities of real-world conditions. This study examines how various environmental factors impact the performance of a hierarchical DRL controller trained for a multi-agent inspection task. Four distinct environments were designed to progressively increase testing fidelity. These are listed below along with primary sources of design complexity:
\begin{enumerate}
    \item \textbf{Training Environment:} A graph-based representation where agents move synchronously and instantly between inspection points. This serves as a means of evaluating baseline policy performance.
    \item \textbf{Hierarchical Inspection Environment:} Simulated agents operating in the Clohessy-Wiltshire (CW) frame under linearized satellite dynamics. This serves as a means of bridging high-level guidance with low-level control to evaluate asynchronicity in policy evaluation and deployment.
    \item \textbf{LINCS Simulation Environment:} Contains high-fidelity satellite state dynamics including support for J2 perturbation and RTA constraints. Provides exposure to unmodeled (during training) dynamic factors that can have influence on policy stability and efficacy. 
    \item \textbf{LINCS Cyber-Physical Testbed:} A hardware-in-the-loop setup where agents are emulated using quadrotor UAVs. Provides policy exposure to physical hardware effects. This includes instance process noise, and latency demonstrating how non-idealities influence the performance of low-level controller.
\end{enumerate}

These environments have a transitory effect on the policy performance. Factors that impact the low-level control directly will reveal themselves over time to the high-level guidance controller. For example, physical trajectory deviations from the idealized graph representation may cause desynchronization between high-level task assignments and real-time execution. The performance of the hierarchical inspection policy is evaluated using the following metrics: completion rate, cumulative distance traveled, and total time taken. These were chosen to align with the reward structure used during training; distance is implicitly minimized by both the high-level and low-level controllers whereas time-management is adjusted through the temporal discount rate set in the training algorithm Proximal Policy Optimization (PPO) algorithm~\cite{schulman2017}. By examining changes in these criteria across environments, this study helps quantify the impact of increasing fidelity and real-world complexities on the hierarchical policy's stability and performance.

 To achieve this, we systematically structure experiments to test the DRL controller's robustness to modeling uncertainties, environmental disturbances, and hardware feedback effects. Such an approach relies on the careful establishment of baseline performance conditions for both the low-level single-agent point-to-point controller and the high-level multi-agent task allocation policy. These benchmarks are derived from training environments where agents operate without \textit{unseen} physical or environmental complexities. Iteratively, we structure three experiments of increasing complexity to evaluate against the baseline. Experiment 1 tests the integration of the high-level guidance with low-level control against deviation with respect to trajectory planning efficiency and task completion. Experiment 2 builds on this by adding the J2 perturbation and RTA intervention into the dynamics and control loop respectively. This assess the sensitivity of our RL control scheme with respect to unmodeled dynamic disturbances. Experiment 3 then evaluates the hierarchical framework in a mixed environment, combining simulated and emulated agents. The performance of the emulated agent, subject to real-world feedback and disturbances, is compared with its simulated counterparts. Table~\ref{tab:experiment_outline} summarizes the experiments and their corresponding key evaluation factors.

\begin{table}[h!]
    \centering
    \resizebox{\textwidth}{!}{%
    \begin{tabular}{|c|c|p{8cm}|}
        \hline
        \textbf{Experiment} & \textbf{Environment} & \textbf{Key Evaluation Factors} \\ \hline
        Baseline: Low-Level (LL) & Training Environment - LL & Performance of the point-to-point controller in isolated low-fidelity environments. \\ \hline
        Baseline: High-Level (HL) & Training Environment - HL & Task allocation and inspection performance in multi-agent setups without physical disturbances. \\ \hline
        Experiment 1 & Hierarchical Inspection Environment & Integration of low- and high-level controllers with trajectory optimization in CWH dynamics. \\ \hline
        Experiment 2 & LINCS Simulation Environment & RTA injection, J2 perturbation, and asynchronous multi-agent operations. \\ \hline
        Experiment 3 & LINCS Cyber-Physical Testbed & Real-world dynamics, sensor feedback, physical disturbances, and mixed simulation-hardware configurations. \\ \hline
    \end{tabular}%
    }
    \caption{Experiment Overview: Environment Details and Evaluation Factors}
    \label{tab:experiment_outline}
\end{table}
\vspace{-10pt}

\FloatBarrier
\section{Experiment Results and Discussion}
\label{section:expr_results}

Results and corresponding discussion are provided below for each experiment in Table~\ref{tab:experiment_outline}. Each baseline experiments is described and included as a point of comparison for Experiment 1. 

\subsection{Experiment 1 - Hierarchical Inspection Environment}

\begin{figure}[!htb]
\begin{subfigure}[b]{1\textwidth}
    \centering
    \hspace*{-1cm}\includegraphics[width=.45\textwidth]{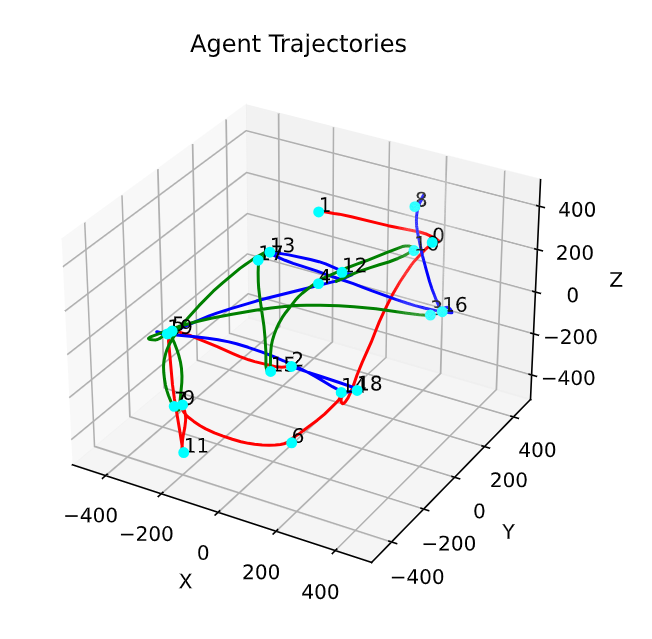}
\end{subfigure}
\begin{subfigure}[b]{0.45\textwidth}
\centering
\includegraphics[width=.85\textwidth]{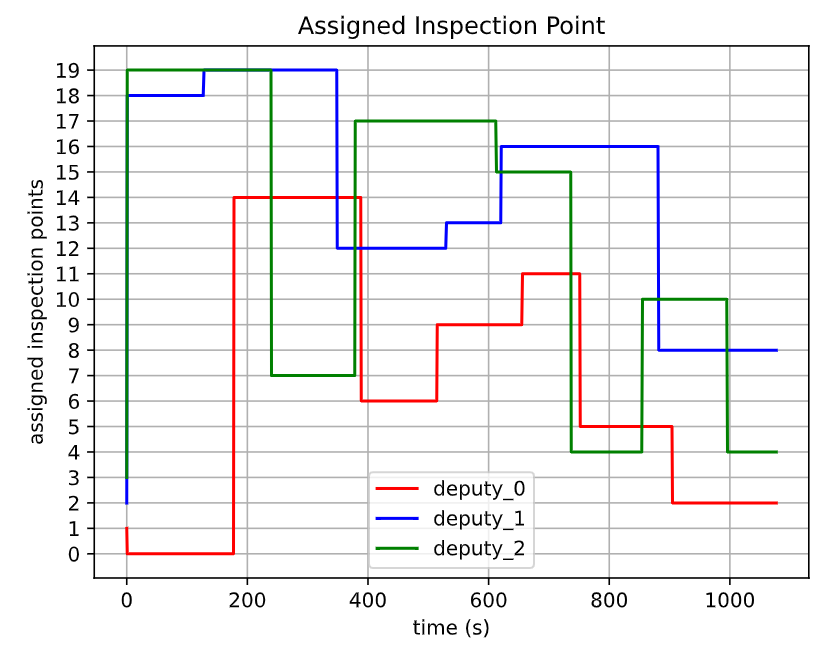}
\end{subfigure}
\begin{subfigure}[b]{0.45\textwidth}
\centering
\includegraphics[width=.85\textwidth]{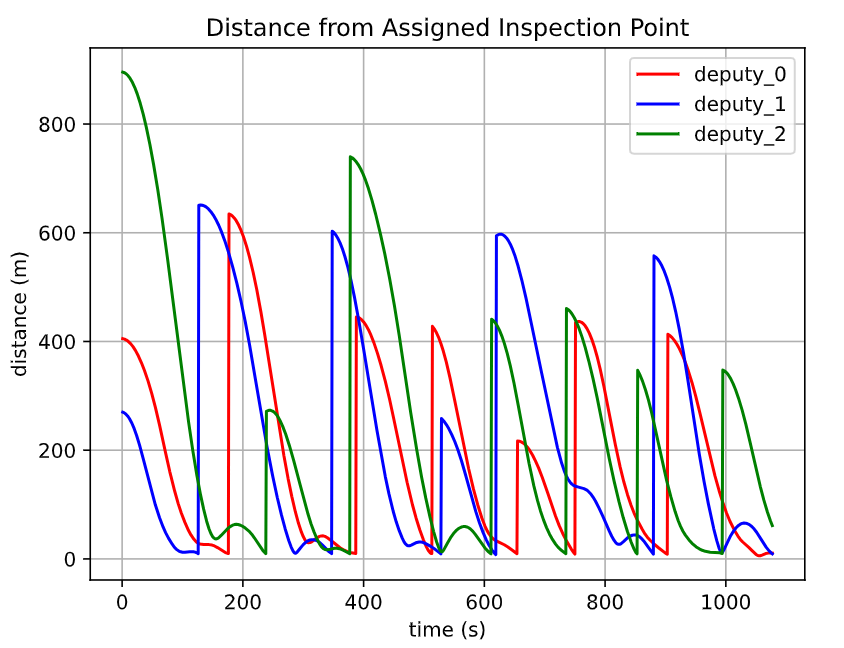}
\end{subfigure}
\caption{Example trajectory for the hierarchical controller as generated in the Hierarchical Inspection Environment. The top figure illustrates trajectories specified through relative position of deputies over time. This is overlayed on the graph of inspection points. The bottom-left figure describes HL guidance specified to the index of inspection point visitation over time. The bottom-right figure illustrates distance from goal as described by the LL controller.}
\label{fig:hier_rollout}
\end{figure}

The hierarchical inspection controller was deployed in a simulated environment to evaluate its ability to integrate high-level guidance and low-level point-to-point control policies. This experiment tested the controller's effectiveness in achieving full coverage of inspection targets while maintaining task efficiency and stability under modeled trajectory uncertainties. The environment consisted of 20 inspection points arranged in a graph-based ellipsoid, with three agents collaboratively performing the inspection task. Key metrics such as time taken, distance traveled, and fuel consumption were measured and compared to a static baseline derived from the high-level and low-level policy baselines. This was constructed to asses the impact that policy inference timing asynchronicity has on task efficacy. The hierarchical controller composed independently of a pretrained guidance planner and point-to-point motion controller consistently achieved full graph coverage, successfully visiting all inspection points across 50 trials. Key metrics considered for evaluation are summarized in Table~\ref{tab:hierarchical_env_results} and trajectory information from a single trial is plotted in Figure~\ref{fig:hier_rollout}

\begin{table}[h!]
    \centering
    \resizebox{0.6\textwidth}{!}{%
    \begin{tabular}{|c|c|c|}
        \hline
        \textbf{Performance Metric} & \textbf{Static Baseline} & \textbf{Hierarchical Sim} \\ \hline
        Targets Reached & 20 & 20 \\ \hline
        Time Taken (s) & 1012.9 & 1105.6 ± 125.7 \\ \hline
        Distance Traveled (m) & 11200.7 & 10720.3 ± 1771.9 \\ \hline
        Straight Line Distance (m) & 7237.2 & 9051.0 ± 1209.2 \\ \hline
        Fuel Consumption (\(\Delta V\)) (m/s) & 355.1 & 382.6 \\ \hline
    \end{tabular}%
    }
    \caption{Key evaluation metrics for Experiment 1. The Static Baseline acts as a comparison statistic extrapolated by the mean performance of HL and LL policies acting independently within their training environments. The Hierarchical Controller is quoted as the sample mean $\pm$ one standard deviation calculated over 50 independent trials.}
    \label{tab:hierarchical_env_results}
\end{table}

The results confirm that the hierarchical inspection controller is capable of cointegrating guidance and control policies to achieve robust multi-agent coordination. While the hierarchical policy showed slight increases in time and fuel consumption compared to the static baseline, it demonstrated significant adaptability to physical trajectory constraints, maintaining stable and efficient performance. These findings validate the controller’s stability and scalability in low-fidelity simulation environments, providing a solid foundation for transitioning to higher-fidelity and hardware-in-the-loop experiments.

\paragraph{Key Observations}
\begin{itemize}
    \item \textbf{Full Coverage:} The hierarchical controller successfully completed the inspection mission in all trials, visiting all 20 inspection points.
    \item \textbf{Time Efficiency:} The hierarchical controller required 9.09\% more time than the static baseline, reflecting the additional time needed for trajectory adjustments in response to low-level dynamics.
    \item \textbf{Distance Efficiency:} Agents under the hierarchical policy traveled 4.29\% less distance than the static baseline, demonstrating the efficacy of the integrated point-to-point controller.
    \item \textbf{Trajectory Complexity:} The hierarchical controller exhibited a 25.06\% increase in cumulative straight-line distance, indicating adjustments in trajectory planning to accommodate physical dynamics.
    \item \textbf{Fuel Consumption:} Fuel usage increased by 7.72\% compared to the baseline, consistent with the added demands of real trajectory execution. It should be noted, that neither controller was trained to optimize over fuel usage. This is an ancillary metric used for stability encapsulation rather than performance analysis.
\end{itemize}

\subsection{Experiment 2 - LINCS Simulation Environment}

\begin{figure}[!htb]
    \begin{subfigure}[b]{0.45\textwidth}
        \includegraphics[width=\textwidth]{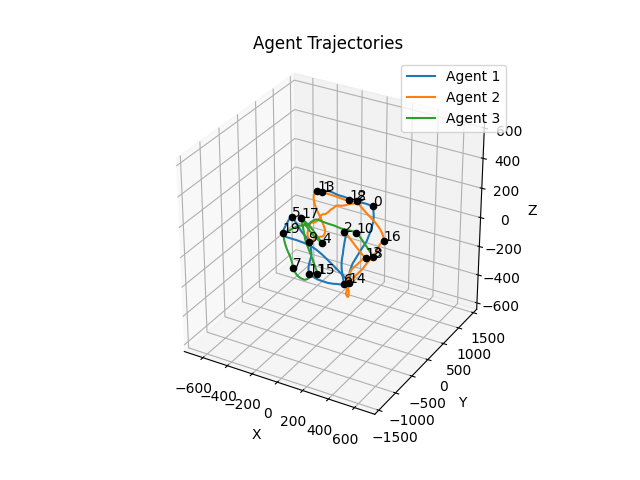}
    \end{subfigure}
    \hfill
    \begin{subfigure}{0.45\textwidth}
        \includegraphics[width=\textwidth]{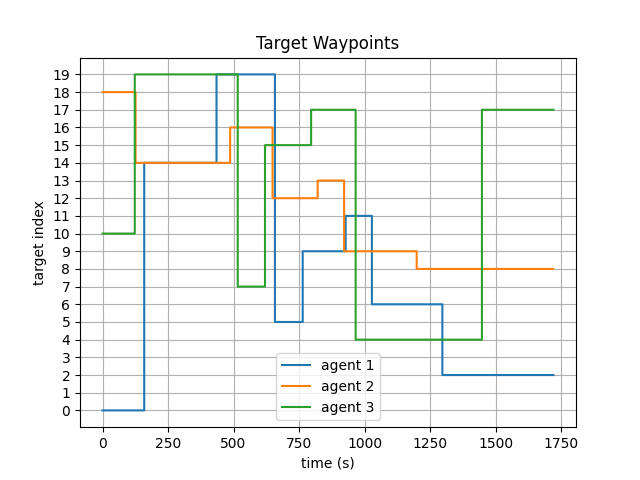}
    \end{subfigure}
    \hfill
    \begin{subfigure}{0.45\textwidth}
        \includegraphics[width=\textwidth]{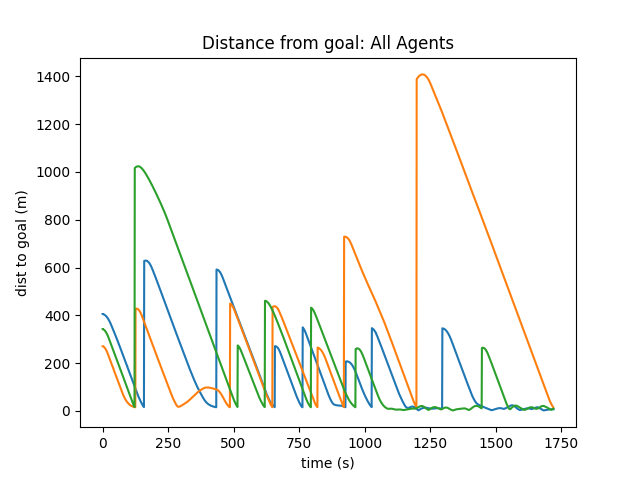}
    \end{subfigure}
    \hfill
    \begin{subfigure}[b]{0.45\textwidth}
        \includegraphics[width=\textwidth]{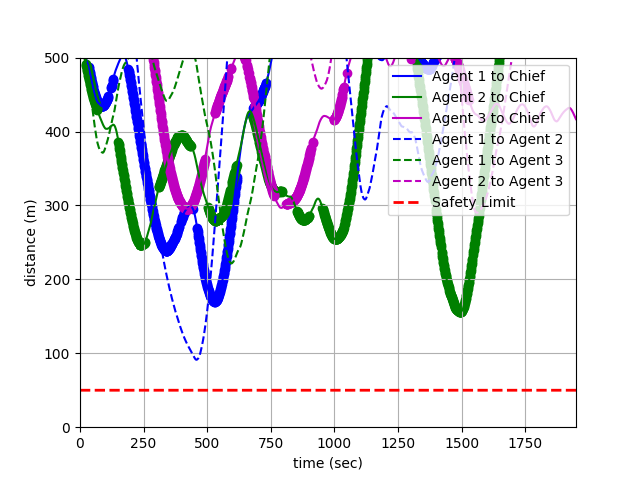}
    \end{subfigure}
    \caption{Trajectory data for Experiment 2. The top-left figure illustrates trajectories specified through relative position of deputies over time. The top-right shows HL guidance specified to the index of inspection point visitation over time. The bottom-left illustrates distance from goal as described by the LL controller. The bottom-right shows RTA activation data over time.}
    \label{fig:expr_3_sim_data}
\end{figure}

The LINCS simulation experiment evaluates the hierarchical controller’s performance under a higher-fidelity environment incorporating J2 perturbations and RTA constraints. The agents operate asynchronously, and their trajectories are influenced by both the added dynamics and the interaction between the guidance and control policies. This experiment provides insights into how the hierarchical framework adapts to dynamic uncertainties and interventions. The agents were initialized at inspection points located at \((-202.7, 433.0, 172.2)\), \((202.7, -433.0, 173.2)\), and \((202.7, 433.0, -173.2)\) with zero initial velocity. The simulation continued until all 20 inspection points were visited by either the assigned agent or by communication from other agents. Four key performance metrics were tracked: total targets reached, time taken, distance traveled, and straight-line distance. This was constructed to asses the impact that higher fidelity state transition dynamics and unanticipated control intervention has on task efficacy when combined with the timing effects of Experiment 1. Therein, the hierarchical controller successfully completed the inspection mission, reaching all 20 targets across all runs. Metrics for evaluation are summarized in Table~\ref{tab:lincs_simulation_results}, with direct comparison to the results of Experiment 1. 

\begin{table}[h!]
    \centering
    \resizebox{0.6\textwidth}{!}{%
    \begin{tabular}{|c|c|c|}
        \hline
        \textbf{Metric} & \textbf{Hierarchical Sim} & \textbf{LINCS Sim} \\ \hline
        Targets Reached & 20 & 20 \\ \hline
        Time Taken (s) & 1105.6 ± 125.7 & 1720.00 (+4.9 sd) \\ \hline
        Distance Traveled (m) & 10720.3 ± 1771.9 & 11472.8 (+0.4 sd) \\ \hline
        Straight Line Distance (m) & 9051.0 ± 1209.2 & 10705.2 (+1.4 sd) \\ \hline
    \end{tabular}%
    }
    \caption{Key evaluation metrics for Experiment 2. A single trial in the LINCS Simulation Environment is conducted and compared against the sample collected from Experiment 1. The single trial draw is expressed both in absolute terms as well as the number of standard deviations in excess of the Hierarchical Simulation mean.}
    \label{tab:lincs_simulation_results}
\end{table}

The results demonstrate that the hierarchical controller remains robust under the added complexities of the LINCS simulation environment at the cost of performance degradation. While there is a noticeable increase in mission duration and trajectory complexity, the framework consistently completes the inspection task without significant performance degradation. The observed deviation in trajectories (see Figure~\ref{fig:expr_3_sim_data} relative to Figure~\ref{fig:hier_rollout}) highlights the importance of enhancing coordination mechanisms and adapting high-level guidance policies to real-time dynamics. As task completion rates are not impacted, these findings support the hypothesis that both HL guidance and LL control are robust to the higher fidelity dynamics in a way that respects the RTA positional constraints. 

\paragraph{Key Observations}
\begin{itemize}
    \item \textbf{Full Coverage:} The hierarchical controller successfully completed the inspection mission, visiting all 20 inspection points despite the added complexity of J2 perturbations and RTA constraints.
    \item \textbf{Increased Time Taken:} The mission duration increased by 55.57\% compared to the hierarchical inspection environment, reflecting the impact of unmodeled dynamics and RTA-induced trajectory adjustments.
    \item \textbf{Distance Efficiency:} Agents traveled 7.01\% more distance in the LINCS simulation, largely due to deviations introduced by dynamic disturbances and asynchronous operations.
    \item \textbf{Straight Line Deviation:} The cumulative straight-line distance increased by 18.31\%, indicating more complex trajectory planning to adapt to the higher-fidelity dynamics.
    \item \textbf{Coordination Challenges:} Instances of "revisits" (points inspected multiple times) were observed, highlighting limitations in task coordination under asynchronous conditions.
\end{itemize}

\subsection{Experiment 3 - LINCS Cyber-Physical (CP) Testbed}

\begin{figure}[!htb]
    \begin{subfigure}[b]{0.45\textwidth}
        \includegraphics[width=\textwidth]{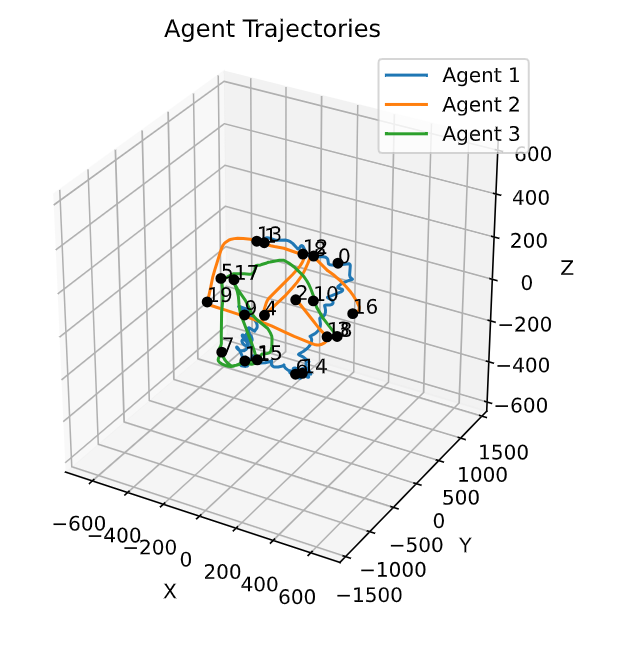}
    \end{subfigure}
    \hfill
    \begin{subfigure}{0.45\textwidth}
        \includegraphics[width=\textwidth]{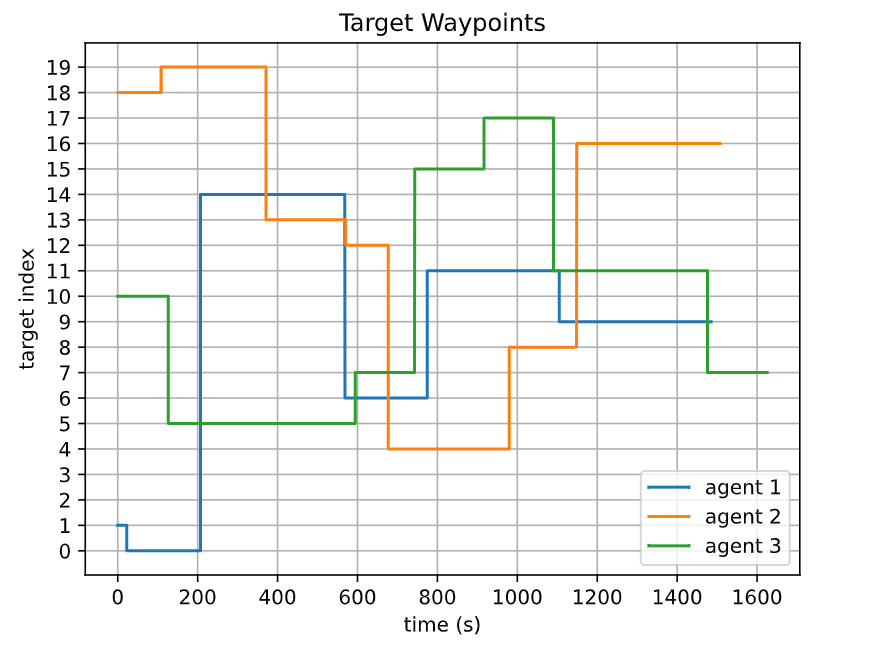}
    \end{subfigure}
    \hfill
    \begin{subfigure}{0.45\textwidth}
        \includegraphics[width=\textwidth]{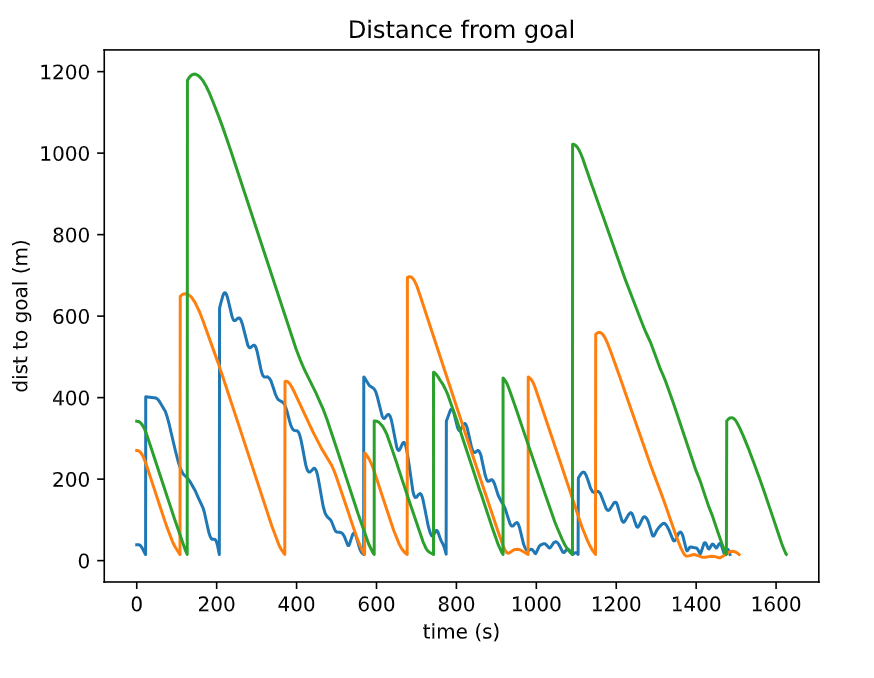}
    \end{subfigure}
    \hfill
    \begin{subfigure}[b]{0.45\textwidth}
        \includegraphics[width=\textwidth]{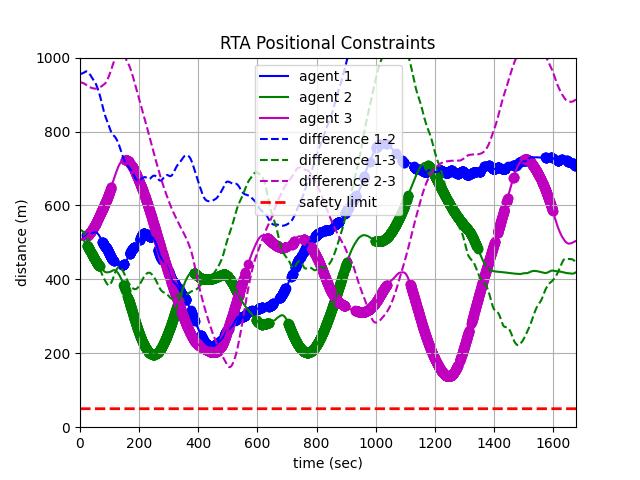}
    \end{subfigure}
    \caption{Trajectory data for Experiment 3. The top-left figure illustrates trajectories specified through relative position of deputies over time. The top-right shows HL guidance specified to the index of inspection point visitation over time. The bottom-left illustrates distance from goal as described by the LL motion controller. The bottom-right shows RTA activation data over time. For this trial, the velocity limit was exceeded multiple times (bolded) while the interagent distance constraint was not triggered.}
    \label{fig:expr_3_data}
\end{figure}

The LINCS cyber-physical testbed experiments evaluate the hierarchical controller’s performance in a CP testbed. This environment introduces real-world challenges, including sensor feedback, dynamic disturbances, and RTA mechanisms, to assess the stability and adaptability of the hierarchical framework. A multi-agent scenario was designed where two agents were simulated and one was emulated by a physical UAV operating in the CP testbed. The three agents are tasked with collaboratively visiting 20 inspection points. Initial position was specified in the same configuration as in Experiment 2 with initial relative velocity pre-stabilized in a neighborhood of 0. This Experiment builds from Experiment 2 through the inclusion of additional physical hardware effects including significant instance process noise and variable timing latency. This experiment indicated that although deputy trajectory characteristics are significantly impacted, task completion rate is not. Given sufficient resources, the DRL controller is still able to reliably complete its mission. Metrics for evaluation are summarized in Table~\ref{tab:multi_agent_results}, with direct comparison to the results of Experiment 1 and 2. 

\begin{table}[h]
    \centering
    \resizebox{0.75\textwidth}{!}{%
    \begin{tabular}{|c|c|c|c|}
        \hline
        \textbf{Metric} & \textbf{Hierarchical Sim} & \textbf{LINCS Sim} & \textbf{LINCS CP Testbed} \\ \hline
        Targets Reached & 20 & 20 &  20 \\ \hline
        Time Taken (s) & 1105.6 (SD=125.7) & 1720.00 (+4.9 sd) & 1484.6 (+3.0 SD) \\ \hline
        Distance Traveled (m) & 10720.3 (SD=1771.9) & 11472.8 (+0.4 sd) & 12961.6 (+1.3 SD) \\ \hline
        Straight Line Distance (m) & 9051.0 (SD=1209.2) & 10705.2 (+1.4 sd)& 9492.1 (+0.4 SD) \\ \hline
    \end{tabular}%
    }
    \caption{Key evaluation metrics for Experiment 3. A single trial in the LINCS CP Testbed is conducted and compared against the sample collected from Experiment 1. The single trial draw is expressed both in absolute terms as well as the number of standard deviations in excess of the Hierarchical Simulation mean.}
    \label{tab:multi_agent_results}
\end{table}

This experiment helps validate the hierarchical controller’s robustness to real-world conditions. While performance metrics such as time taken and distance traveled increased due to environmental disturbances, the inspection missions were completed without critical failures. These findings demonstrate the stability of the hierarchical framework when transitioning from a simulated environment to a CP testbed, underscoring its potential for real-world applications in autonomous satellite operations. The inclusion of the emulated agents appears to have an impact on RTA activation, specifically with respect to the constraint imposed on relative velocity - see Figure~\ref{fig:expr_3_data}. The activation frequency of the emulated agent appears slightly less dense, possible due to the small back-and-forth oscillations induced by the quadrotor tracking induced perturbations. The inclusion of the emulated agents appears to have an impact on RTA activation, specifically with respect to the constraint imposed on relative velocity - see Figure~\ref{fig:expr_3_data} as compared to Figure~\ref{fig:expr_3_sim_data}. This could be due to scaled noise emanating from the emulated agent creating spikes in relative velocity.

\paragraph{Key Observations}
\begin{itemize}
    \item \textbf{Task Completion:}
    The hierarchical controller successfully completed the inspection mission, visiting all 20 inspection points across the three agents. However, the time taken increased by 34.27\%, primarily due to RTA interventions and asynchronous dynamics in the testbed. Additionally, the distance traveled increased by 17.29\%, reflecting the influence of physical agent dynamics on trajectory efficiency. 
    \item \textbf{Runtime Assurance Effects:} RTA interventions slowed down trajectories by limiting velocities and preventing collisions, which distorted point-to-point paths. As compared against the LINCS simulation results, time taken actually decreases while total distance traveled increased. This supports a tendency for agents to move at a higher relative velocity requiring more frequent intervention by the RTA. Despite the significant intervention rate, the controller was still able to accomplish its objectives.
    \item \textbf{Physical Feedback Impact:} The UAV in the testbed exhibited trajectory deviations due to atmospheric effects, requiring oscillations to stabilize around targets. These oscillations when scaled back to Hill's frame create impule spikes in velocity that are appropriately smoothed. This results in a tendency to move more quickly than in both \textit{pure} simulation environments.
\end{itemize}

\section{Conclusion}
\label{section:conclusion}
This study evaluated the stability and performance of a hierarchical DRL framework for multi-agent satellite inspection tasks. The framework integrates a high-level guidance policy with a low-level motion controller to address key challenges in autonomous satellite operations, including task allocation, trajectory optimization, and adaptation to dynamic uncertainties. Through a comprehensive series of experiments across simulation and cyber-physical environments, the hierarchical controller demonstrated consistent performance, achieving high task completion rates while maintaining robustness under varying environmental conditions. By progressively increasing fidelity and complexity, the experiments highlighted the framework's ability to bridge the sim-to-real gap and adapt to real-world disturbances such as sensor feedback, dynamic perturbations, and RTA interventions.

Key performance metrics, including completion rate, time taken, and distance traveled were analyzed to characterize the effects of environmental complexities on the controller. The results showed that while higher-fidelity environments introduced additional challenges, the hierarchical controller successfully maintained stability and completed the inspection tasks without critical failures. The LINCS CP testbed experiments particularly underscored the impact of physical dynamics and feedback delays, validating the framework's scalability and adaptability to real-world conditions. Overall, this study provides evidence of the hierarchical DRL framework's potential for enabling robust and efficient autonomous satellite operations in increasingly complex and dynamic environments.

\section*{ACKNOWLEDGMENT}
S. Phillips was funded directly by the Air Force Research Laboratory (AFRL) as Research Staff. H. Lei, Z. Lippay, A. Zaman, J. Aurand, and A. Maghareh were funded by AFRL under FA9453-21-C-0602.

\bibliography{references}
\end{document}